\documentclass{article}

\usepackage[preprint]{neurips_2025}

\usepackage[utf8]{inputenc} 
\usepackage[T1]{fontenc}    
\usepackage{hyperref}       
\usepackage{url}            
\usepackage{booktabs}       
\usepackage{amsmath}        
\usepackage{amsfonts}       
\usepackage{nicefrac}       
\usepackage{microtype}      
\usepackage[table]{xcolor}  

\usepackage{graphicx}
\usepackage{booktabs}
\usepackage{adjustbox}
\usepackage{siunitx}
\usepackage{caption}

\definecolor{bestblue}{RGB}{0,45,114}
\newcommand{\best}[1]{\textbf{\textcolor{bestblue}{#1}}}

\newcommand{\ours}{Maglev}

\usepackage{tikz}
\usetikzlibrary{arrows.meta,backgrounds}

\title{\ours{}: Sliding Recurrent Memory}

\author{%
  Bo Liu, Qiang Liu\\
  The University of Texas at Austin\\
  \texttt{\{bliu, lqiang\}@cs.utexas.edu} \\
}

\begin{document}

\maketitle

\begin{abstract}
We introduce \ours{}, a recurrent Transformer architecture with fixed-size memory that generalizes sliding-window attention while remaining parallelizable during training. \ours{} consists of two coupled models: a prefiller $Q$, which leverages full attention\footnote{In practice, we use interleaved full and sliding-window attention for $Q$, as this yields stronger performance. The essential requirement is that $Q$ be more expressive than $P$, with access to the full history.} to produce memory targets $m'_t$, and a decoder $P$, which uses only sliding-window attention and recurrent K/V injection to produce decoder memories $m_t$ for next-token prediction. We train \ours{} with a memory consistency loss that aligns $m_t$ with $m'_t$, allowing inference to use $P$ alone. Empirically, \ours{} improves validation loss and downstream pretraining benchmarks over sliding-window and latent recurrent transformer baselines. Moreover, sharing parameters between $P$ and $Q$ reduces parameter memory while preserving most of the gains.
\begin{figure}[h!]
    \centering
    \vspace{-5pt}
    \includegraphics[width=0.78\linewidth]{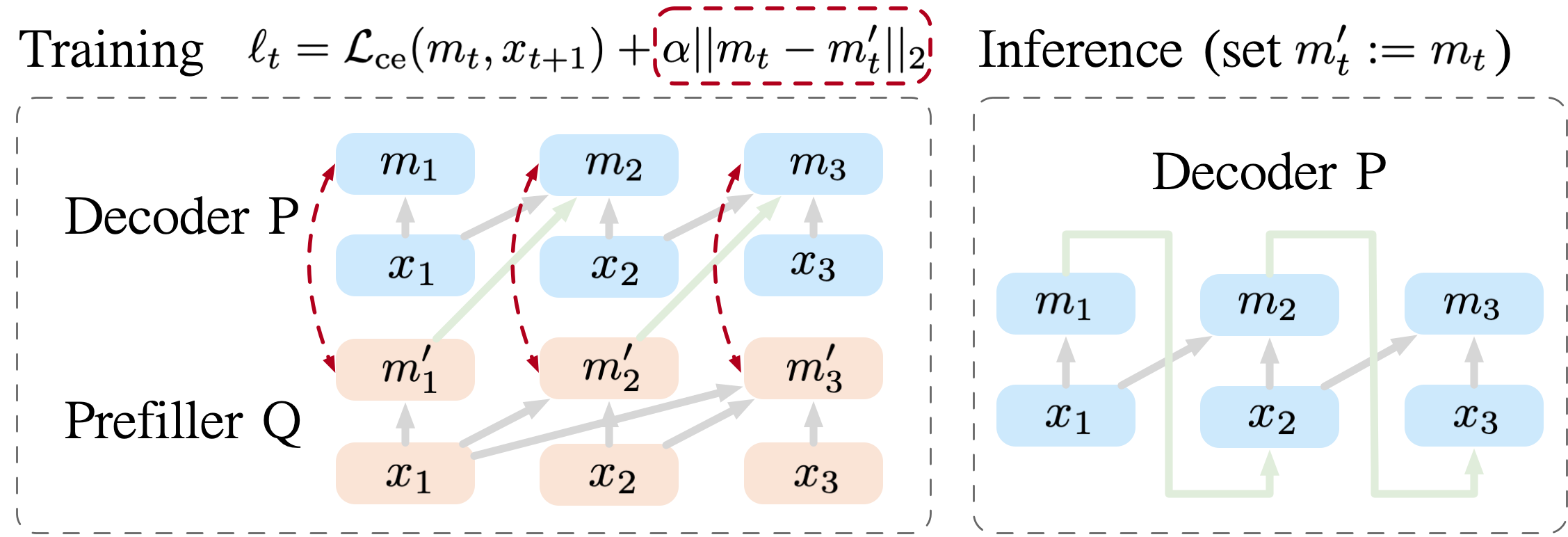}
    \captionsetup{width=0.8\linewidth}
    \caption{
    \ours{}: a prefiller $Q$ produces memory targets $m'_t$ from the observed sequence, and a decoder $P$ consumes the shifted $m'_{t-1}$ while predicting $x_{t+1}$ and producing its own memory $m_t$. The consistency loss aligns $m_t$ with $m'_t$. At inference, the prefiller is discarded and the decoder runs recurrently using its own memories. In practice, we find that we could largely share parameters between $P$ and $Q$ without degrading much performance.}
    \label{fig:header}
\end{figure}
\end{abstract}

\section{Introduction}
\label{sec::intro}
Transformers have a simple memory strategy: keep every token. Full causal self-attention lets each prediction revisit the entire prefix, enabling powerful nonlinear retrieval, copying, and in-context computation \citep{vaswani2017attention}; its cost, however, grows with context. Sliding-window attention bounds both attention and KV-cache costs, but discards distant information and therefore provides no persistent long-term memory \citep{beltagy2020longformer,jiang2023mistral}. Classical recurrent models such as LSTMs compress history with nonlinear token-wise updates, but propagate a relatively small state sequentially, limiting both capacity and sequence parallelism \citep{hochreiter1997lstm}. Memory Transformers offer richer states, yet commonly update them only at segment or block boundaries, imposing an artificial granularity on memory \citep{dai2019transformerxl,rae_compressive_2019,hutchins2022block,bulatov2022rmt}. Linear attention and state-space models remove these boundaries and parallelize token-wise recurrence, but do so through structured linear or affine updates \citep{katharopoulos2020transformers,s4,gu_mamba_2023,pmlr-v235-dao24a,Liu2024LonghornSS,yang2024gateddelta}. The missing piece is a practical way to give a nonlinear Transformer persistent memory that it can rewrite after every token.

\paragraph{Goal: token-wise recurrent memory with parallel pretraining.}
We study language models with \emph{token-wise latent memory}. After processing $x_t$, the model writes a latent memory vector $m_t$ and retains the bounded window
\begin{equation}
\mathcal{M}_t=\{m_{t-W+1},\ldots,m_t\},
\end{equation}
ignoring boundary effects. Writing all trainable parameters as $\Theta$, the desired recurrent interface predicts the next token and writes the current memory from only a recent token window and the preceding memory window,
\begin{equation}
p_\Theta(x_{t+1},m_t\mid x_{t-W+1:t},\mathcal{M}_{t-1}).
\end{equation}
A practical token-wise memory model should combine three properties:
\begin{itemize}
    \item \textbf{Token-wise nonlinear memory.} Unlike the structured, layer-local linear or affine updates of linear attention and state-space models \citep{katharopoulos2020transformers,s4,gu_mamba_2023,pmlr-v235-dao24a}, each memory update traverses the full nonlinear Transformer depth, providing looped-Transformer-like recurrent depth without extra Transformer iterations \citep{dehghani2019universal,giannou2023looped}. Because every token's decoder state serves as memory through its ordinary K/V entry, the recurrence also requires no dedicated memory tokens or cache and introduces no artificial segment or block boundaries \citep{dai2019transformerxl,rae_compressive_2019,hutchins2022block,bulatov2022rmt}.
    \item \textbf{Bounded recurrent inference.} The token and memory windows remain fixed as the sequence grows. As in sliding-window attention \citep{beltagy2020longformer,jiang2023mistral}, inference uses a bounded local KV cache; the recurrent memories carry information beyond that window without increasing its size.
    \item \textbf{Parallel large-scale pretraining.} Although inference is recurrent, ideally pretraining should process all positions in parallel, preserving standard Transformer throughput \citep{vaswani2017attention}. Achieving this parallelism is nontrivial for nonlinear recurrence \citep{lim2024parallelizingnonlinearsequentialmodels,gonzalez2024towards}.
\end{itemize}
The interface provides the first two properties by construction: it writes $m_t$ at every position while keeping both conditioning windows fixed. Parallel pretraining, however, is not automatic. Text provides $x_{1:T}$ but not the latent trajectory $m_{1:T}$; because each update consumes preceding memories, direct training must generate this trajectory sequentially. The central question is how to preserve the recurrent interface at deployment without unrolling it during pretraining.

\paragraph{Maglev: parallel supervision for recurrent memory.}
\ours{} breaks this dependency by separating memory construction during pretraining from memory propagation at inference. A stronger causal prefiller $Q$ produces target memories $m'_{1:T}$ from the observed sequence in parallel. A sliding-window decoder $P$ consumes the shifted targets $m'_{0:T-1}$, predicts the next tokens, and emits memories $m_{1:T}$. A consistency loss aligns each $m_t$ with $m'_t$, teaching $P$ to produce the memory needed at the next step. Training therefore requires two sequence-parallel passes. At inference, $Q$ is removed and $P$ closes the loop with its own memories.

Figure~\ref{fig:header} illustrates the architecture. The deployed decoder remains close to an ordinary sliding-window Transformer, augmenting the same local attention pattern with a gated K/V pathway for shifted recurrent memory. The decoder's final normalized state is both the memory $m_t$ and the input to the language-modeling head. Fixed token and memory windows keep inference costs bounded.

\paragraph{Contributions.}
We propose a token-wise nonlinear recurrent-memory Transformer that folds shifted memories into sliding-window K/V entries, preserving bounded inference without extra memory tokens. We introduce lifted parallel training, where a causal prefiller $Q$ generates targets in parallel and consistency training teaches decoder $P$ to reproduce them without sequential recurrent unrolling. With a $435$M-scale deployed decoder trained on $43.52$B tokens, our best model improves FineWeb-Edu validation BPB from $0.7413$ to $0.7251$ and average downstream accuracy from $54.1$ to $56.4$ over a matched sliding-window Transformer, while also outperforming matched latent recurrent Transformer. Moreover, $Q$ and $P$ can share most parameters while preserving most gains.

\section{Method}
\label{sec::method}

Maglev replaces sequential latent-state training with two sequence-parallel Transformer passes. As summarized in Figure~\ref{fig:header}, a causal prefiller $Q$ constructs an auxiliary memory trajectory, and a deployable decoder $P$ learns to use and reproduce it one step at a time. At inference, $Q$ is discarded and $P$ feeds back its own memories, recovering token-wise recurrence. We first formalize this construction and then describe our Transformer instantiation.

\subsection{Lifted Parallel Training}

Direct recurrent training would generate $m_1,m_2,\ldots,m_T$ in order. \emph{Lifted parallel training} instead constructs an explicit auxiliary trajectory from the observed sequence in one causal, sequence-parallel pass. The prefiller $Q$ may have a larger receptive field than the decoder, but remains causal so that its memories contain no future information.

Let $x_{1:T}$ be a token sequence and $d$ the hidden dimension. Let $\phi$ and $\theta$ denote the parameters of $Q$ and $P$, respectively, and let $\Theta=\phi\cup\theta$ denote all unique trainable parameters. The sets $\phi$ and $\theta$ may overlap. Training consists of
\begin{equation}
\underbrace{\textcolor{bestblue}{\boldsymbol{m}'_{1:T}}=Q_\phi(x_{1:T})}_{\text{prefiller pass}},
\qquad
\underbrace{\textcolor{bestblue}{\boldsymbol{m}_{1:T}}=P_\theta(x_{1:T},\textcolor{bestblue}{\boldsymbol{m}'_{0:T-1}})}_{\text{decoder pass}},
\label{eq:lifted-passes}
\end{equation}
where $m'_0=0$. Here $m'_t$ and $m_t$ are the prefiller and decoder outputs after all Transformer blocks and the final RMS normalization. At position $t$, the shifted input gives $P$ a memory window ending at $m'_{t-1}$, from which it produces $m_t$. Because $Q$ is causal, this window depends only on $x_{\leq t-1}$ and matches the information available at inference. Despite the sequence notation in Equation~\ref{eq:lifted-passes}, causal sliding-window attention restricts each position to its local token context and corresponding shifted-memory window.

During training, $P$ receives the shifted prefiller trajectory $m'_{0:T-1}$; during inference, it receives its own preceding memories $m_{0:T-1}$ through the same channel. Because $Q$ constructs the complete target trajectory in one causal parallel pass, all shifted targets are available before the decoder pass, allowing $P$ to evaluate every position simultaneously. Maglev thus replaces a length-$T$ unroll with two sequence-parallel training passes while retaining recurrent inference.

Only the decoder predicts tokens. Its final memory $m_t$ is also the state read by the language-modeling head:
\begin{equation}
o_t = U m_t,
\qquad
p_\Theta(x_{t+1}\mid x_{\leq t})=\operatorname{softmax}(15\tanh(o_t/15)).
\label{eq:decoder-prediction}
\end{equation}
For valid target positions $\mathcal I$, let $y_t=x_{t+1}$. We optimize
\begin{equation}
\mathcal L(\Theta)
= \frac{1}{|\mathcal I|}\sum_{t\in\mathcal I}
\bigg[\underbrace{
\operatorname{CE}\bigl(p_\Theta(\cdot\mid x_{\leq t}),y_t\bigr)
}_{\mathcal L_{\rm CE}}
+
\lambda
\underbrace{
\frac{\lVert m_t-m'_t\rVert_2}{\sqrt d}
}_{\mathcal L_{\rm cons}} \bigg].
\label{eq:maglev-objective}
\end{equation}
The first term trains next-token prediction from decoder memory $m_t$. The second aligns $m_t$ with prefiller target $m'_t$. From a shifted prefiller-memory window, $P$ learns both to predict the next token and to produce the memory for the next window. This consistency enables recurrent inference using decoder memories in place of prefiller memories.

\subsection{Transformer Instantiation}

\paragraph{Prefiller, decoder, and parameter sharing.}
The prefiller is a training mechanism rather than a prescribed architecture: any causal model that produces all $m'_t$ in parallel can serve as $Q$. Our $Q$ and $P$ are Transformers following the nanochat layer-pattern convention \citep{karpathy2025nanochat}, in which a pattern lists successive attention types and repeats through the stack. For example, \texttt{SLSL} alternates sliding-window (\texttt{S}) and full causal (\texttt{L}) attention. The prefiller uses \texttt{SLSL}, with a $512$-token sliding window and full attention over the $2048$-token training context. The decoder uses \texttt{SSSS}, restricting every layer to $W=512$. Thus, $Q$ has more context for constructing targets, while $P$ retains the bounded receptive field required for deployment.

By default, $Q$ and $P$ share Transformer blocks, with separate residual-scaling parameters for the decoder path. We also evaluate a variant with a separate $Q$ stack. Both use the same two-pass objective in Equations~\ref{eq:lifted-passes}--\ref{eq:maglev-objective}; only the block sharing differs.

\paragraph{Recurrent K/V injection.}
Following the latent recurrent Transformer \citep{huang2026latent}, the decoder injects shifted memory through K/V features rather than dedicated memory tokens. At decoder layer $\ell$, let $a_t^\ell$ be the incoming residual stream and $q_t^\ell,k_t^\ell,v_t^\ell$ the local query, key, and value. During training, shared projections map the shifted prefiller memory to recurrent K/V features:
\begin{equation}
k_t^{\rm rec}=\operatorname{RMSNorm}(W_k^{\rm rec} m'_{t-1}),
\qquad
v_t^{\rm rec}=W_v^{\rm rec} m'_{t-1}.
\end{equation}
At inference, decoder memory $m_{t-1}$ replaces $m'_{t-1}$. Layer-specific gates mix local and recurrent features,
\begin{equation}
g_{\rm loc,t}^\ell=2\sigma(G_{\rm loc}^\ell a_t^\ell),
\qquad
g_{\rm rec,t}^\ell=2\sigma(G_{\rm rec}^\ell a_t^\ell),
\end{equation}
\begin{equation}
\bar k_t^\ell=g_{\rm loc,t}^\ell\odot k_t^\ell+g_{\rm rec,t}^\ell\odot k_t^{\rm rec},
\qquad
\bar v_t^\ell=g_{\rm loc,t}^\ell\odot v_t^\ell+g_{\rm rec,t}^\ell\odot v_t^{\rm rec}.
\end{equation}
The factor of two sets each gate to the neutral value one at zero pre-activation, while allowing independent scaling of the two channels. Sliding-window attention operates on the mixed K/V cache,
\begin{equation}
c_t^\ell=
\operatorname{Attention}\!\left(
q_t^\ell,
\bar k_{\max(1,t-W+1):t}^\ell,
\bar v_{\max(1,t-W+1):t}^\ell
\right).
\label{eq:recurrent-attention}
\end{equation}
Each cached entry at position $j$ therefore combines features of local token $x_j$ and shifted memory $m'_{j-1}$ during training, or $m_{j-1}$ during inference. Attention over the last $W$ mixed entries exposes the corresponding memory window without allocating additional sequence positions. When enabled, RoPE is applied to both local and recurrent keys before mixing. After all Transformer blocks and the final normalization, the resulting decoder state is $m_t$; it serves both as the next recurrent memory and as the input to the language-modeling head.


\subsection{Inference}

At inference, $Q$ is discarded and $P$ runs recurrently,
\begin{equation}
\underbrace{\textcolor{bestblue}{\boldsymbol{m}_t}
=P_\theta(x_{t-W+1:t},\textcolor{bestblue}{\mathcal M_{t-1}})}_{\text{recurrent inference}},
\qquad
\textcolor{bestblue}{\mathcal M_{t-1}}=\{\textcolor{bestblue}{m_{t-W}},\ldots,\textcolor{bestblue}{m_{t-1}}\}.
\label{eq:recurrent-inference}
\end{equation}
The language-modeling head uses $m_t$ to predict $x_{t+1}$, and $m_t$ is appended to the recurrent window. The $W$ existing K/V entries represent both token and memory histories, so Maglev matches the cache size and attention cost of ordinary sliding-window attention, independent of sequence length.
\section{Related Work}
\label{sec::related}

\paragraph{Efficient and long-context attention.}
Transformers rely on softmax attention over past tokens \citep{vaswani2017attention}. Long-context variants reduce this cost with local attention, compressed memory, recurrence, retrieval, or hybrid memory mechanisms. Transformer-XL and Compressive Transformers reuse or compress segment-level activations \citep{dai2019transformerxl,rae_compressive_2019}; Infini-attention and landmark-style methods add compressed or random-access memory for longer contexts \citep{munkhdalai2024leave,mohtashami2023landmark}; Memorizing Transformers augment attention with external retrieval \citep{wu2022memorizing}. Sliding-window attention is especially practical because it bounds compute and cache size, but it drops information outside the window. \ours{} keeps the sliding window and adds a shifted memory that carries information forward through the decoder.

\paragraph{Linear recurrent and convolutional sequence models.}
Linear attention and state space models replace the full key-value cache with a fixed-size recurrent state \citep{katharopoulos2020transformers,s4,gu_mamba_2023,pmlr-v235-dao24a}. Related convolutional and state-space models such as H3, S5, Hyena, and Mamba use efficient recurrent or convolutional sequence operators \citep{fu2022hungry,smith2022simplified,poli2023hyena,gu_mamba_2023}. Wavelet-inspired multiresolution convolutional memory instead summarizes the history at exponentially increasing temporal scales using learned filters shared across a tree of dilated causal convolutions \citep{shi2023sequence}. More recent gated recurrent architectures, including RetNet, RWKV, Griffin/Hawk, gated delta networks, and xLSTM, add data-dependent gates or learned update rules for stronger memory control \citep{sun2023retentive,peng2023rwkv,de_griffin_2024,yang2024gateddelta,beck2024xlstm}. These approaches are efficient, but their memory transformations are typically linear, affine, or specialized recurrent or convolutional operators. \ours{} also uses a bounded state, but its state is produced by a nonlinear Transformer decoder and trained through prefiller consistency; unlike multiresolution convolutional memory, its state size does not grow with the number of temporal scales.

\paragraph{Hybrid attention and recurrence.}
Hybrid models combine local attention with recurrent or state-space layers so recent tokens remain easy to access while older context is compressed \citep{de_griffin_2024,ren2024samba,lieber2024jamba}. Feedback Transformers and TransformerFAM expose high-level past representations to future computation \citep{fan2021feedback,hwang2024transformerfam}; Recurrent Memory Transformers and Block-Recurrent Transformers carry memory tokens or block states across segments \citep{bulatov2022rmt,hutchins2022block}; block-state and retention-style models balance recurrence and parallelism through blockwise computation \citep{pilault2023block,sun2023retentive}. \ours{} shares the goal of bounded memory, but keeps the deployed model close to a sliding-window Transformer by injecting a shifted memory through the decoder's K/V pathway.

\paragraph{Online and test-time memory.}
Several recent works view sequence models as online learners or test-time memory systems \citep{longhorn,ttt,behrouz2024titanslearningmemorizetest}. Other work studies how to parallelize nonlinear recurrent computation \citep{lim2024parallelizingnonlinearsequentialmodels,gonzalez2024towards}. \ours{} is closest in spirit to this line because its recurrent update is nonlinear, but it avoids sequential training by learning from a parallel prefiller rather than by directly unrolling the decoder recurrence.

\paragraph{Latent recurrence and extra computation.}
Depth-recurrent and latent-thinking methods add computation by looping blocks or inserting auxiliary tokens before prediction \citep{dehghani2019universal,giannou2023looped,geiping2026scaling,goyal2024think,pfau2024let,herel2024thinking,zelikman2024quiet,hao2024training}. Concurrently, LRT~\citep{huang2026latent} passes a previous-token memory state into the next token and trains this recurrence through parallel refinement passes. Other concurrent work also uses parallel supervision for recurrent memory: Supervised Memory Training \citep{kumar2026pretraining} trains general RNNs from Transformer-generated memory labels, while Rec2PM \citep{chen2026recurrentpreference} trains compact preference-memory updates for long-sequence generative recommendation. Maglev applies a similar idea to language-modeling Transformers: it injects shifted token memories through the K/V pathway, retaining the attention-window and cache profile of ordinary sliding-window attention while carrying a richer recurrent state. It also supports sharing parameters between the prefiller $Q$ and decoder $P$, tying the parallel training signal directly to the recurrent Transformer rather than using a fully separate teacher--student system.

\section{Experiments}
\label{sec::experiment}

We evaluate Maglev in the nanochat pretraining stack~\citep{karpathy2025nanochat}. All models use the same tokenizer, data pipeline, optimizer family, sequence length, and evaluation scripts; architecture-specific differences are described below.

\paragraph{Training setup.}
We use the d20 nanochat architecture: $L=20$ layers, width $d=1280$, head dimension $128$, $10$ attention heads, and maximum sequence length $2048$. The short-window size is $W=512$. The standard d20 model has $435{,}159{,}040$ non-embedding scaling parameters, and we train for $43.52$B tokens, corresponding to a $100\times$ Chinchilla-style token budget when embedding parameters are excluded~\citep{hoffmann2022training}. The optimizer batch contains $524{,}288$ tokens per step, and we use the MuonAdamW training recipe~\citep{jordan2024muon}.

\paragraph{Models.}
We compare against an interleaved full/sliding-window Transformer with layer pattern \texttt{SLSL}, a purely sliding-window model with layer pattern \texttt{SSSS}, and LRT variants using the same two layer patterns. Both LRT and Maglev use the same shared recurrent K/V injection and residual/input-skip update. In Maglev, the prefiller $Q$ uses \texttt{SLSL} and the decoder $P$ uses \texttt{SSSS}. We evaluate shared-parameter Maglev and a separate-parameter variant, each with consistency weight $\lambda\in\{0.1,1.0\}$.

\paragraph{Evaluation.}
We report FineWeb-Edu validation bits per byte (FW BPB)~\citep{lozhkovfineweb}, LAMBADA perplexity and accuracy~\citep{paperno2016lambada}, and common pretraining downstream benchmarks: PIQA~\citep{bisk2020piqa}, HellaSwag~\citep{zellers2019hellaswag}, WinoGrande~\citep{sakaguchi2021winogrande}, ARC-Easy and ARC-Challenge~\citep{clark2018think}, SocialIQA~\citep{sap2019socialiqa}, and BoolQ~\citep{clark2019boolq}. The average column is the mean of the reported downstream accuracies.

\begin{figure}[t!]
    \centering
    \includegraphics[width=\linewidth]{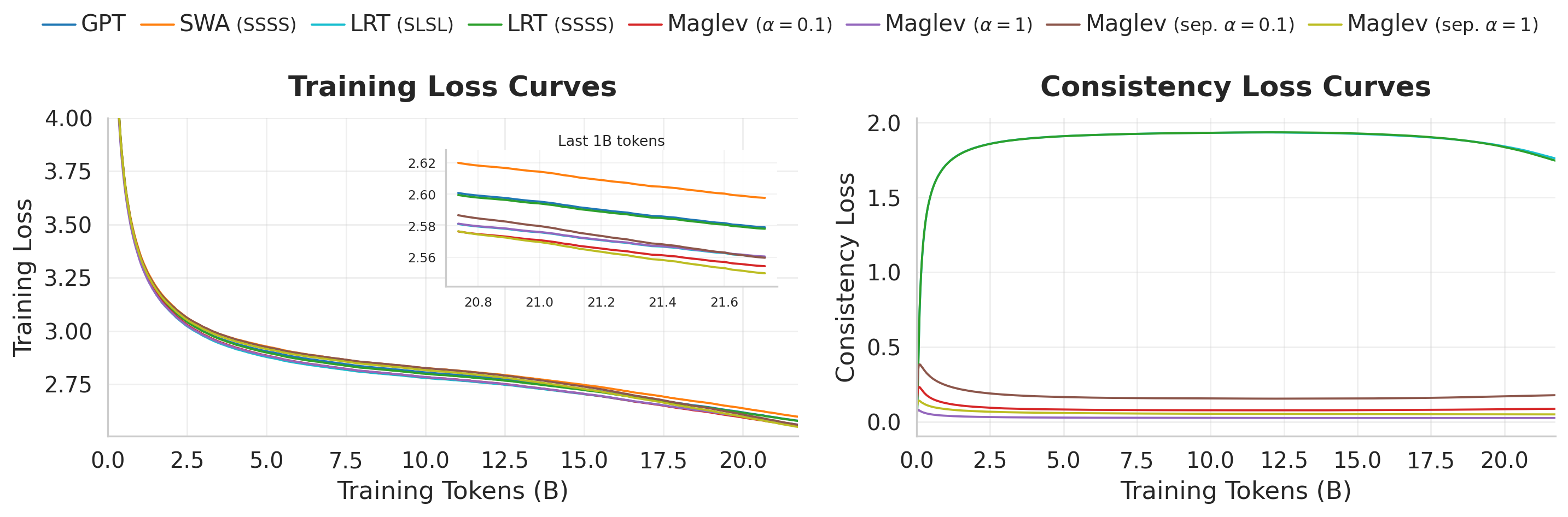}
    \caption{Training dynamics for the Maglev consistency objective. The prefiller $Q$ supplies memory targets $m'_t$, and the decoder $P$ is trained to produce matching memories $m_t$ while optimizing next-token prediction.}
    \label{fig:training_consistency}
\end{figure}

\begin{table}[t]
\centering
\small
\setlength{\tabcolsep}{4.5pt}
\renewcommand{\arraystretch}{1.08}

\begin{adjustbox}{width=\linewidth}
\begin{tabular}{
    l
    S[table-format=1.4, table-number-alignment=center, table-text-alignment=center]
    S[table-format=2.2, table-number-alignment=center, table-text-alignment=right]
    | S[table-format=2.1, table-number-alignment=center, table-text-alignment=center]
    S[table-format=2.1, table-number-alignment=center, table-text-alignment=center]
    S[table-format=2.1, table-number-alignment=center, table-text-alignment=center]
    S[table-format=2.1, table-number-alignment=center, table-text-alignment=center]
    S[table-format=2.1, table-number-alignment=center, table-text-alignment=center]
    S[table-format=2.1, table-number-alignment=center, table-text-alignment=center]
    S[table-format=2.1, table-number-alignment=center, table-text-alignment=center]
    S[table-format=2.1, table-number-alignment=center, table-text-alignment=center]
    | S[table-format=2.1, table-number-alignment=center, table-text-alignment=center]
}
\toprule
\rowcolor{gray!10}
\textbf{Model}
& {\textbf{FW}}
& {\textbf{LMD}}
& {\textbf{LMD}}
& {\textbf{PIQA}}
& {\textbf{Hella}}
& {\textbf{Wino}}
& {\textbf{ARC-E}}
& {\textbf{ARC-C}}
& {\textbf{SIQA}}
& {\textbf{BoolQ}}
& {\textbf{Avg}} \\

\rowcolor{gray!10}
&
{\scriptsize \textbf{BPB} $\downarrow$}
& {\scriptsize \textbf{PPL} $\downarrow$}
& {\scriptsize \textbf{Acc} $\uparrow$}
& {\scriptsize \textbf{Acc} $\uparrow$}
& {\scriptsize \textbf{Acc$_n$} $\uparrow$}
& {\scriptsize \textbf{Acc} $\uparrow$}
& {\scriptsize \textbf{Acc} $\uparrow$}
& {\scriptsize \textbf{Acc$_n$} $\uparrow$}
& {\scriptsize \textbf{Acc} $\uparrow$}
& {\scriptsize \textbf{Acc} $\uparrow$}
& {\scriptsize \textbf{Acc} $\uparrow$} \\
\midrule

Transformer (SLSL) & 0.7373 & 8.44 & 45.9 & \best{73.7} & 53.0 & 57.7 & 68.6 & 38.8 & 41.7 & 56.9 & 54.5 \\
LRT (SLSL) & 0.7292 & 7.94 & 47.7 & 72.4 & 54.9 & 58.1 & 70.1 & 39.4 & 40.7 & 63.6 & 55.9 \\
\midrule
SWA (SSSS) & 0.7413 & 8.54 & 46.2 & 70.5 & 53.3 & 57.1 & 68.4 & 40.1 & 41.3 & 56.2 & 54.1 \\
LRT (SSSS) & 0.7331 & 7.92 & 47.3 & 72.3 & 54.9 & \best{58.8} & \best{70.4} & 39.8 & 40.7 & 56.2 & 55.0 \\
\midrule
Maglev ($\lambda=0.1$) & 0.7295 & 8.06 & 47.3 & 72.0 & \best{55.2} & 58.7 & 70.0 & 41.3 & 41.6 & 63.6 & 56.2 \\
Maglev ($\lambda=1$) & 0.7320 & 8.27 & 46.3 & 72.4 & 53.8 & 56.4 & 69.7 & 39.8 & 41.2 & 51.1 & 53.9 \\
Maglev (sep. $\lambda=0.1$) & 0.7276 & \best{7.73}& \best{48.7} & 72.2 & 54.7 & 57.9 & 69.2 & 40.5 & \best{42.7} & 62.0 & 56.0 \\
Maglev (sep. $\lambda=1$) & \best{0.7251} & 8.06 & 47.4 & 72.6 & 55.2 & 57.4 & 69.7 & \best{42.7} & 42.1 & \best{64.0} & \best{56.4} \\
\bottomrule
\end{tabular}
\end{adjustbox}

\vspace{2pt}
\caption{
Pretraining benchmark results after $43.52$B training tokens. FW BPB is evaluated on FineWeb-Edu validation data; LMD denotes LAMBADA. HellaSwag and ARC-Challenge use normalized accuracy. Higher is better except for BPB and perplexity.
}
\label{tab:gb200-eval}
\end{table}

\paragraph{Results.}
Maglev improves the fixed-window decoder without relying on full attention at inference. The shared Maglev model with $\lambda=0.1$ reaches $0.7295$ FW BPB and $56.2$ average downstream accuracy, improving over both the SSSS sliding-window baseline and the corresponding LRT baseline. The separate-parameter variant with $\lambda=1$ gives the best FW BPB ($0.7251$) and average downstream score ($56.4$). The comparison suggests that additional prefiller capacity can improve the consistency target, while the shared model uses less parameter memory and keeps the two paths tightly coupled.

The results also show that the consistency weight is not purely monotone. A larger $\lambda$ improves the separate-parameter model, but hurts the shared model on several downstream tasks. This is consistent with the role of $Q$: when $Q$ and $P$ share most parameters, an overly strong memory-consistency term can constrain the decoder representation; when $Q$ has separate capacity, the stronger target can provide a more useful training signal.
\section{Conclusion and Future Work}
\label{sec::conclusion}
We introduced \ours{}, a fixed-memory recurrent Transformer trained through a prefiller--decoder consistency objective. The prefiller provides parallel memory targets $m'_t$, while the decoder learns to predict tokens and produce its own memories $m_t$ using sliding-window attention with recurrent K/V injection. At inference, the prefiller is removed and the decoder runs as a bounded-memory recurrent model. Our experiments show that this scheme improves validation BPB and downstream pretraining benchmarks over sliding-window and LRT baselines in the nanochat d20 setting. More broadly, Maglev provides a way to train a nonlinear recurrent model with fixed inference memory while preserving parallel training.

Because our experiments were constrained by available compute, Maglev remains a preliminary investigation rather than a definitive study of this design space. Several directions remain open. First, scaling Maglev will require studying the tradeoff between prefiller strength and decoder capacity, together with kernels that make recurrent injection efficient in deployment. Second, $Q$ need not be trained from scratch with $P$: a pretrained or lightly fine-tuned language model could provide memory targets while only $P$ is trained, distilling its representations into a compact recurrent decoder. Third, our results show that substantial sharing between $Q$ and $P$ is possible, but the best sharing pattern remains unclear; intermediate designs could share embeddings, MLPs, attention projections, or selected layers while retaining task-specific components. Finally, recurrent K/V injection is only one way to expose past memories. Future work should compare alternatives such as residual-stream injection, recurrent tokens, cross-attention, and layer-specific memory projections, characterizing their tradeoffs in expressivity, stability, parameter cost, cache size, and inference throughput.

\bibliographystyle{plainnat}
\bibliography{neurips_2025}

\end{document}